%% file: root.tex
\documentclass[letterpaper, 10 pt, conference]{ieeeconf}  
\IEEEoverridecommandlockouts   
\usepackage{url}

\usepackage[all]{nowidow}

\usepackage{pgfplots}
\pgfplotsset{compat=1.17}

\usepackage{algorithm2e}
\usepackage{math}

\usepackage{booktabs}

\usepackage{threeparttable}

\newcommand\mypara[1]{\vspace{1mm}\noindent\textbf{#1}}

\title{\LARGE \bf WGICP: Differentiable Weighted GICP-Based Lidar Odometry}

\author{Sanghyun Son, Jing Liang, Ming Lin, Dinesh Manocha \\
Project page: \url{https://gamma.umd.edu/researchdirections/autonomousdriving/diff_odom}
}

\begin{document}

\maketitle
\thispagestyle{empty}
\pagestyle{empty}

\begin{abstract} 
We present a novel differentiable weighted generalized iterative closest point (WGICP) method applicable to general 3D point cloud data, including that from lidar. Our method builds on differentiable generalized ICP (GICP) and we use the differentiable K-Nearest Neighbors (KNN) algorithm to enhance differentiability. Our differentiable GICP algorithm provides the gradient of output pose estimation with respect to each input point. This allows us to train a neural network to predict its importance, or weight, in order to estimate the correct pose. In contrast to the other ICP-based methods, our formulation reduces the number of points used for GICP by only selecting those with the highest weights and ignoring redundant ones with lower weights. We show that our method improves both the accuracy (up to 30\%) and the speed (up to 2$\times$) over the GICP algorithm on the KITTI dataset. We also demonstrate the benefit of WGICP in terms of developing an improved SLAM system.  

\end{abstract}

\input{1_introduction}
\input{2_related_work}

\input{3_preliminaries}
\input{4_method}
\input{5_experiments}
\input{6_conclusion}

\section{Acknowledge}
This work was supported in part by ARO Grants W911NF2110026,  and Army Cooperative Agreement W911NF2120076.

\bibliographystyle{IEEEtran}
\bibliography{references}


\end{document}

%% file: 1_introduction.tex
\section{INTRODUCTION}
\label{sec:introduction}

3D Simultaneous Localization and Mapping (SLAM) methods are used to solve the problems  of localization and map building in unknown scenarios~\cite{slam2006, park2018elastic}. In this context,  localization refers to  estimating the position and orientation of the mobile robot in 3D space. Given the estimated pose, mapping functions are used to model the environment using the perceived information.

The localization step is usually performed by odometry algorithms~\cite{forster2016manifold, zhang2014loam, shan2018lego, buchanan2022learning,dellenbach2021s, liosam,wang2021f,zhang2017low, quenzel2021real}. These odometry algorithms use sensory inputs from the latest consecutive frames to estimate the relative pose of a mobile robot with respect to the global frame while the robot is possibly moving. If we use a lidar as the main perception device, the sensory inputs correspond to 3D point clouds. In terms of lidar odometry, we can define the overall goal as estimating the optimal transformation matrix that best aligns the given point cloud input to the global frame. Since each frame of a point cloud is captured in real-time (e.g., 40Hz for Hokuyo UTM-30LX Laser Scanner~\cite{zhang2014loam}), we usually compare only the latest two frames of input for efficiency~\cite{zhang2014loam, li2019net}. In many cases, the overall accuracy can be further improved by using other sensory inputs (e.g., IMU)~\cite{shan2018lego, liosam}, but the underlying accuracy of the odometry itself is still important. 

Many algorithms have been proposed to solve the lidar odometry problem. Among them, rule-based methods are still widely used because of their real-time performance~\cite{liosam}. ICP-based methods~\cite{besl1992method, segal2009generalized} improve the pose estimate in an iterative manner.  Despite their simplicity and generalizability, these methods may not work well for the cases where the input point cloud has a non-uniform distribution or is sparse~\cite{grant2013finding, li2019net}. In addition, when the number of points increase, it results in increasing the computational cost. To alleviate these issues, voxel-based down-sampling or matching methods can be used to regularize the non-uniform points with voxels or replace the costly K-Nearest Neighbors (KNN) algorithm~\cite{biber2003normal, koide2021voxelized}. However, it results in a tradeoff between the accuracy and speed.




Recently, learning-based methods have been proposed to solve the odometry problem. These methods can provide promising results in terms of improving the accuracy, as compared to rule-based methods~\cite{velas, li2019net, buchanan2022learning}. However, these approaches mainly use trained neural networks to directly map input point clouds to the transformation matrix in the output. Since they do not rely on an existing model-based algorithm, we need a lot of point cloud training data to train them.

\noindent {\bf Main Results:}
 We present a novel learning-based method for lidar odometry. Our method is based on the {\em differentiable} generalized iterative closest point (GICP) algorithm. We chose GICP because it is one of the most effective extensions of ICP and computes the optimal alignment by assuming local planar distributions in the point clouds. Our differentiable GICP algorithm is based on the differentiable ICP algorithm~\cite{jatavallabhula2019gradslam}. To enhance the differentiability, we use the differentiable KNN algorithm in our method. Based on this differentiable formulation, we also propose a weighted GICP (WGICP) method that can solve the sparsity and speed issues as voxel-based methods while achieving better accuracy with acceleration.
Our novel results include:

\begin{itemize}
    \item {\em Differentiable GICP}: 
    We propose a differentiable GICP algorithm that shows better differentiability than~\cite{jatavallabhula2019gradslam}. With the gradient information, we can compute how each point contributes to the pose estimation in the GICP algorithm. Moreover, it allows us to embed a neural network that could be trained to improve the performance of the GICP method, instead of training a new system from scratch as is the case with other learning methods.
    
    
    
    \item {\em Weighted GICP}: Our differentiable GICP can be integrated with neural networks. We present a neural architecture that learns to predict the \textit{weight}, or \textit{importance}, of each point while performing GICP. We refer to our method as {\em weighted GICP}, as our method uses these per-point weights. In our formulation, the points with lower weights are designed to influence the GICP result less than those with higher weights. Therefore, at runtime, we can reject points with lower weights because they do not contribute to the correct estimation very much. In this way, we can improve both accuracy and speed of the algorithm, which is important for applications of odometry to SLAM.

\end{itemize}

We show that our differential WGICP method offers improved performance over the GICP algorithm for lidar odometry, in terms of better pose estimation and  faster speed. In practice, our neural network  takes less than $5ms$ on NVidia RTX A5000 GPU in terms of processing 10K points during each frame.
Moreover, it allows us to reduce the number of points used for GICP, which accelerates the performance. 
We demonstrate the benefits of our approach on the KITTI dataset~\cite{Geiger2013IJRR}, where we observe up to $30\%$ of improvement in accuracy and up to $2\times$ speedup than the baseline GICP. 
We also show our method can be used for general SLAM tasks with faster and more accurate map building.

%% file: 2_related_work.tex
\section{Related Work}
\label{sec:related_work}

\subsection{Rule-Based Odometry}

The Iterative Closest Point (ICP) method~\cite{besl1992method,rusinkiewicz2001efficient} is one of the most widely used registration methods for 3D point clouds, and many variants have been proposed. Generalized ICP (GICP)~\cite{segal2009generalized}  is one such variant that extends the original algorithm with distribution-based matching and results in better accuracy. These ICP-based registration methods can provide accurate results for general inputs but may not work well for sparse inputs~\cite{grant2013finding, li2019net}. Also, they use every data point for matching since there is no specific selection mechanism, which means that their computational complexity is proportional to the number of points. To relieve these issues, voxel-based downsampling or matching method~\cite{biber2003normal,dellenbach2022ct, koide2021voxelized, quenzel2021real} has been proposed to improve GICP performance. 

Instead of using every point, feature-based methods use a smaller number of points that have geometric features such as planar or edge features~\cite{grant2013finding, velas2016collar}. These techniques have been widely used in lidar-based SLAM algorithms~\cite{zhang2014loam,deschaud2018imls, shan2018lego, liosam, pan2021mulls} as they achieve both real-time speed and accuracy. However, these methods are weak to noisy points that are generated from dynamic objects in the scene as they cannot discriminate them from solid feature points. Therefore, they need a mapping procedure based on dense matching methods like ICP to compensate for accuracy. In addition, most of their feature extraction operations are discrete and do not admit smooth approximation.

\subsection{Learning-Based Odometry}

Learning-based odometry methods for lidar point cloud inputs mainly rely on an image-based approach~\cite{velas, li2019net}. They project an input 3D point cloud onto a 2D rectangular image that is fed into a CNN-based system, which is used to estimate the pose. Since their estimation process is based on neural networks, they need a lot of point cloud data to train for generalization. Also, they cannot guarantee a certain level of accuracy for the inputs they have not seen before. We show that our method achieves better accuracy for unseen data in the KITTI dataset, as our method is based on GICP.


Recently, differentiable SLAM methods have been also proposed~\cite{jatavallabhula2019gradslam, karkus2021differentiable}. Differentiable programming methods are often used to make a computation model differentiable for optimization and control; when it comes to SLAM, instead of training neural networks from scratch, we can train them to improve the existing SLAM algorithm's performance. Even though those methods focused on visual SLAM, Murty et al.~\cite{jatavallabhula2019gradslam} proposed a differentiable ICP algorithm that is based on the differentiable Levenberg-Marquardt (LM) optimization algorithm. However, its differentiability is limited by the KNN algorithm. In this work, we propose using the differentiable KNN algorithm for our differentiable GICP to overcome the limit. Also, while the previous differentiable SLAM methods only focused on differentiability, we propose integrating a neural network to the differentiable algorithm to improve its actual performance.

%% file: 3_preliminaries.tex
\section{Notation \& Problem Formulation}

\begin{figure}[t]
\centering
    \includegraphics[width=0.9\linewidth]{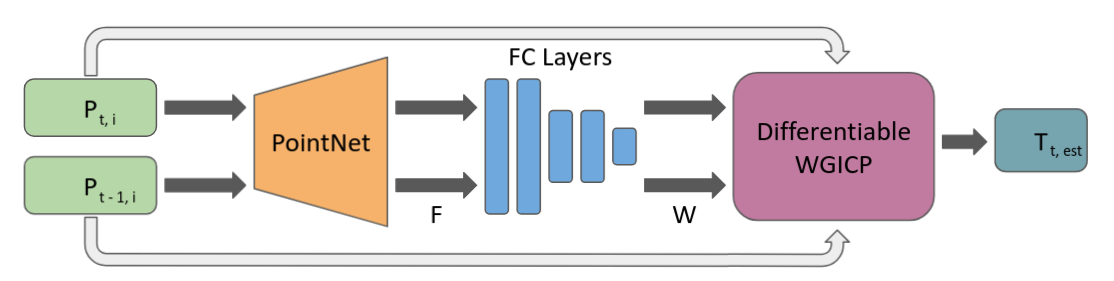}
    \caption{\textbf{Overall pipeline of our differentiable WGICP algorithm.} Given two point clouds, $P_{t-1}$ and $P_{t}$, from two consecutive frames, we first compute per-point feature vectors ($F_{t-1, i}, F_{t, i}$) using PointNet~\cite{qi2017pointnet}. Those feature vectors are forwarded to a fully connected network, which infers per-point weights. Using those weights and input point clouds, our differentiable WGICP algorithm estimates the optimal transformation matrix $\mathbf{T}_{t, est}$ that aligns $P_{t}$ to $P_{t-1}$. To train the embedded neural networks, we  compare $\mathbf{T}_{t, est}$ to ground truth $\mathbf{T}_{t, gt}$ and use back-propagation.}
    \label{fig:pipeline}
    \vspace{-1em}
\end{figure}

\subsection{Notation}
Our approach is designed for a Lidar sensor that is mounted on a mobile robot and provides a 3D point cloud input at every frame. Let us denote the incoming point cloud at frame $t$ as $P_{t}$. We use two major coordinate systems to represent the point clouds. The first one is lidar local coordinate system $L$, which changes over time as the robot moves around.  We denote the lidar local coordinate system at time $t$ as $L_{t}$. At the same time, there is a world coordinate system, $W$, where we want to place all the points and use them to build the map. We use the lidar local coordinate system of the first frame, $L_{1}$, as $W$, following the convention of previous SLAM algorithms~\cite{zhang2014loam}. Additionally, we denote the $i$-th point in $P_{t}$ as  $P_{t, i}$; it will be described as $P_{t, i}^{j}$ in $L_{j}$, and as $P_{t, i}^{W}$ in $W$. 

The relative pose, or transformation of the mobile robot, at time $t$ is denoted as $\mathbf{T}_{t}$, and it is used to transform a point in $L_{t}$ to $W$. Specifically, $\mathbf{T}_{t}$ can be represented with a $4\times4$ matrix, which is used to compute the world coordinates as
\begin{equation*}
    P_{t, i}^{W} = \mathbf{T}_{t} P_{t, i}^{t}. 
\end{equation*}

\subsection{Problem Definition}
The main task of the odometry algorithm is to align two point clouds from consecutive frames; that is, given a point cloud $P_{t}$, we want to find the optimal transformation $\widetilde{\mathbf{T}_{t}}$ that brings a point $P_{t, i}^{t}$ to its corresponding location in $L_{t-1}$. We can describe this problem formally as follows:
\begin{equation*}
    \text{Find the transformation } \widetilde{\mathbf{T}_{t}} \text{ s.t. }\widetilde{\mathbf{T}_{t}}  P_{t, i}^{t} = P_{t, i}^{t-1}. 
\end{equation*}

ICP-based methods solve this problem by minimizing the sum of distances between matching points. Please refer to Section~\ref{sec:approach} for details. Note that we can use this transformation matrix $\widetilde{\mathbf{T}_{t}}$ for computing $\mathbf{T}_{t}$ as
\begin{equation*}
    \mathbf{T}_{t} = \mathbf{T}_{t-1}\widetilde{\mathbf{T}_{t}},
\end{equation*}
which can be used for transforming the points in the local coordinates $L_{t}$ to the global coordinates $W$ in the mapping step.

%% file: 4_method.tex
\section{Our Approach}
\label{sec:approach}




\input{4-1_odometry}


%% file: 4-1_odometry.tex
Based on GICP, we present a {\em differentiable} extension that adaptively learns weights for the network. For two perceived point clouds, we first use PointNet~\cite{qi2017pointnet} as the backbone to generate weights of the perceived points. Next, the differentiable GICP is applied to the weighted points to estimate the relative pose of the Lidar sensor (Figure~\ref{fig:pipeline}).

\subsection{Differentiable GICP}
\label{sec:diff-gicp}

In this section, we briefly introduce GICP and present our method to differentiate it. Our method is based on the differentiable ICP algorithm~\cite{jatavallabhula2019gradslam}, and we use the differentiable KNN algorithm~\cite{DBLP:journals/corr/abs-1810-12575} to improve the overall differentiability. We aim to compute reliable gradient information with our differentiable GICP method, so we can use it for our WGICP algorithm described in Section~\ref{sec:wgicp}.

\subsubsection{GICP}

For given source point cloud $A$ and target point cloud $B$, the ICP algorithm finds the optimal transformation matrix $\mathbf{T}\ast$ that aligns $A$ to $B$ by minimizing the sum of distances between matching points. For each point $a_{i}$ in $A$, we can find the matching point $b_{i}$ in $B$ with the KNN algorithm. After matching, we redefine the set of matching points $B=\{b_i\}_{i=1, \dots, N}$ as a corresponding order of $A=\{a_i\}_{i=1, \dots, N}$. Define the ICP optimization problem as:

\begin{equation}
    \label{eq:icp}
    \mathbf{T}\ast = arg\min_{\mathbf{T}} \sum_{i} {d_{i}^{(\mathbf{T})}}^{T}d_{i}^{({\mathbf{T}})}, \text{ where } d_{i}^{(\mathbf{T})} = b_i - \mathbf{T}a_i.
\end{equation}
 ICP only considers positional information of each point. In contrast, GICP also considers distributional information of each point, which is represented with a covariance matrix $C$~\cite{segal2009generalized}. To be specific, GICP assumes that the points are probabilistically generated from the underlying set of points, $\hat{A}=\{\hat{a}_{i}\}$ and $\hat{B}=\{\hat{b}_{i}\}$, according to the normal distribution as follows.
\begin{align*}
    a_i \sim N(\hat{a}_{i}, C^{A}_{i}), b_i \sim N(\hat{b}_{i}, C^{B}_{i}).
\end{align*}
We can compute covariance matrix $C$ for each point by using principal component analysis (PCA) with its nearest neighboring points~\cite{segal2009generalized}.

Next, we can formulate the optimization problem of GICP as
\begin{align}
\label{eq:gicp}
    \mathbf{T}\ast = arg\min_{\mathbf{T}} \sum_{i} {d_{i}^{(\mathbf{T})}}^{T}{(C_{i}^{B} + \mathbf{T}C_{i}^{A}\mathbf{T}^{T})}^{-1}d_{i}^{({\mathbf{T}})}.
\end{align}
Since this objective function is a non-linear function, we often rely on non-linear least squares optimization methods to solve the problem. Similar to other   GICP implementations, we use the Levenberg-Marquardt (LM) algorithm~\cite{bjorck1996numerical}.

\subsubsection{Differentiation}

There are two main non-differentiable algorithms included in the original GICP algorithm. One is the KNN algorithm for finding point correspondences, and the other is the LM algorithm for non-linear optimization.

\mypara{K-Nearest Neighbors:} 
The KNN algorithm is non-differentiable because it computes point correspondences by only considering the single nearest point of the query point. 
Since the gradient only flows through this nearest point, rather than flowing through all the nearby points, the back-propagation step becomes unstable. To address this issue, we can sample $K_{d} > 1$ number of nearest points ($P_{j} |_{j=1, ..., K_{d}}$) for each query point ($P_{i}$) and compute the KNN weights ($w_{i,j}^{knn}$) based on their distances ($||d_{i, j}||_{2} = || P_{i} - P_{j} ||_2$) as:
\begin{equation}
    w_{i, j}^{knn} = w^{knn}(d_{i, j}) = \frac{e^{-||d_{i, j}||_{2}}}{\sum_{j=1, ..., K_{d}} e^{-||d_{i, j}||_{2}}}.
\label{eq:knn_weight}
\end{equation}
Note that this is the special case of the differentiable KNN algorithm in \cite{DBLP:journals/corr/abs-1810-12575}, which has been used for 3D points. We changed the objective term in Equation~\ref{eq:gicp} as follows to accommodate these weights.

\begin{equation}
    \sum_{i} \sum_{j} w_{i, j}^{knn} {d_{i, j}^{(\mathbf{T})}}^{T}{(C_{j}^{B} + \mathbf{T}C_{i}^{A}\mathbf{T}^{T})}^{-1}d_{i, j}^{({\mathbf{T}})}.
\label{eq:knn_weight_loss}
\end{equation}

\mypara{Levenberg–Marquardt(LM):} We can define our non-linear optimization problem as $\min_{x} L(x)$,
where $L$ is an objective function. For current $x_{t}$, we can use either a first order gradient descent method or a second order Gauss-Newton method to compute $\Delta x_{t}$, which can be used to update $x_{t}$ as
\begin{equation*}
    x_{t+1} = x_{t} + \Delta x_{t}.
\end{equation*}
However, in many cases we cannot choose which method is better than another in advance, because the problems are  high-dimensional, non-linear problems.

The LM algorithm addresses this problem by using a damping variable $\lambda > 0$ to find a balance between two methods. If $\lambda$ is large, it takes the safer approach of gradient descent, which could converge slowly. In contrast, when $\lambda$ is small, it takes an unstable but fast update step with the Gauss-Newton method. Based on these properties, the LM algorithm compares the current objective $L(x_t)$ with the objective after one step of update $L(x_t + \Delta x_t)$.  If the objective after the update is smaller than the current one, we can perform the update step and make $\lambda$ smaller to accelerate the performance. Otherwise, we do not take the update step and only increase $\lambda$. 

To differentiate this discrete update scheme, we use the smooth gating functions proposed in~\cite{jatavallabhula2019gradslam}: 
\begin{align}
    \lambda = \lambda_{min} + \frac{\lambda_{max} - \lambda_{min}}{1 + e^{-(L(x_t + \Delta x_t) - L(x_t))}},
\label{eq:diff-lm-lambda} \\
    x_{t+1} = x_{t} + \frac{\Delta x_{t}}{1 + e^{-(L(x_t + \Delta x_t) - L(x_t))}},
\label{eq:diff-lm-transform}
\end{align}
where $\lambda_{min}$ and $\lambda_{max}$ are predefined minimum and maximum values for $\lambda$, respectively.

\RestyleAlgo{ruled}
\begin{algorithm}[t]
\caption{Differentiable WGICP Algorithm}
\label{alg:diff-wgicp}
\KwData{Point clouds $A$ and $B$, $K_{d} > 1$ for KNN. }
\KwResult{$\mathbf{T}\ast$}
$Loss \gets \infty$\;
$\lambda \gets \lambda_{0}$\;
$\textbf{T} \gets I$\;

\textit{// Use networks to infer weights for WGICP}\;
$w_i, w_j \gets $ point weight of $a_i \in A$ and $b_j \in B$\;
\While{$\textbf{T}$ not converged}{
    \textit{// Differentiable KNN algorithm}\;
    Find $K_{d}$ nearest neighbors, $b_{i, j=1,..., K_d}$, for each point $a_{i} \in A$ under current $\textbf{T}$\;
    
    Compute weights $\bar{w}_{knn, j}$ for each neighbors using Equation~\ref{eq:weighted-knn-weight}\;
    
    \
    \textit{// Differentiable LM algorithm}\;
    Compute overall loss $L$ based on Equation~\ref{eq:weighted-gicp-loss}, and get gradient $\Delta \mathbf{T}$ of it using current $\lambda$\;
    
    Compute lookahead loss $L'$ from $\mathbf{T} + \Delta \mathbf{T}$\;
    
    Udpate $\lambda$ and $\mathbf{T}$ using Equation~\ref{eq:diff-lm-lambda},~\ref{eq:diff-lm-transform}\;
}
return $\mathbf{T}$\;
\end{algorithm}


\input{5-1-table}

\subsection{Weighted GICP}
\label{sec:wgicp}
In this section, we present the weighted GICP (Algorithm~\ref{alg:diff-wgicp}), which uses per-point weights as part of the GICP algorithm. Intuitively, this method gives higher weights to points that contribute to the correct pose estimation in the modified GICP algorithm and lower weights to points that do not. These outlier points can include nonuniform, sparse points or those from dynamic objects that are occluded or not contribute towards pose estimation.

Our WGICP method is based on a system of neural networks, which is comprised of a feature extraction network and a fully connected network. We use PointNet~\cite{qi2017pointnet} as the feature extraction network because it is one of the fastest models for processing point clouds. As shown in Figure~\ref{fig:pipeline}, the feature extraction network extracts a per-point feature vector, $F_{i}$, for each point $P_{i}$. Next, the fully connected network computes the per-point weight $0 \le w_{i} \le 1$ based on the feature vector.

If we use these point weights for both the source and target point clouds, $A$ and $B$, we can extend the optimization problem in Equation~\ref{eq:knn_weight_loss} as
\begin{align}
\label{eq:weighted-gicp-loss}
    \sum_{i} w_i \sum_{j} (& \bar{w}_{i, j}^{knn}) {d_{i, j}^{(\mathbf{T})}}^{T}{(C_{j}^{B} + \mathbf{T}C_{i}^{A}\mathbf{T}^{T})}^{-1}d_{i, j}^{({\mathbf{T}})}, \\
    & \text{where } \bar{w}_{i, j}^{knn} = w^{knn}(\frac{d_{i, j}}{w_j}).
\label{eq:weighted-knn-weight}
\end{align}
Note that we divide $d_{i, j}$ by $w_j$ and then compute the KNN weight to preserve the property that KNN weights sum to 1 (Equation~\ref{eq:knn_weight}).
In this formulation, when $w_{i}$ is very small, the point $A_{i}$ does not contribute to the loss or the pose estimation. This is same for $w_{j}$, as its corresponding KNN weight decreases. Therefore, it enables our network to control the weights $w_i$  that are used to compute the appropriate points to consider in GICP algorithm.

\subsubsection{Rejecting Outlier Points}
In this section, we present our method to reject outlier points based on the per-point weights inferred from our model. This is a simple mechanism that  excludes points with lower weights, and run baseline GICP with remaining points. However, to train our model to maximize the performance under this mechanism, we need a differentiable way to to give near-zero weights to the outliers that would be rejected, while we give weights near 1 to the points that would remain. 
We use the following criteria to approximate this mechanism in a differentiable manner:

\begin{equation}
    w_{i}' = sigmoid(\frac{w_i - \mu (w_i)}{\sigma (w_i)}),
\end{equation}
where $\mu (w_i)$ is average weight, and $\sigma (w_i)$ is standard deviation of the weights in a single point cloud.



\subsubsection{Training}
To train our network to predict per-point weights suited for Lidar inputs, we can use two point clouds, $P_{t}$ and $P_{t - 1}$, to our pipeline (Figure~\ref{fig:pipeline}). When we compute the estimated transformation matrix $T_{t, est}$ for those inputs, we can compare it with ground truth transformation matrix $T_{t, gt}$ to compute the estimation error. We use the Frobenius norm of the difference between the transformation matrices to measure the error, or loss, and train our network to minimize this loss.
\begin{align}
    L(T_{t, est}, T_{t, gt}) = ||T_{t, est} - T_{t, gt}||_{\mathbf{F}}.
    \label{eq:loss}
\end{align}


%% file: 5-1-table.tex
\begin{table*}[t]
\centering 
\caption{{\bf Odometry errors on KITTI dataset}. The dataset is divided into two, where 00$\sim$06 sequences correspond to the  training dataset, and 07$\sim$10 sequences are the test dataset. We used PCL~\cite{Rusu_ICRA2011_PCL} implementation for ICP, and the implementation of Koide et al.~\cite{koide2021voxelized} for GICP and VGICP. Since Velas et al.~\cite{velas} and LO-Net~\cite{li2019net} are not publicly available, we directly quoted the performance numbers from those papers, and for LO-Net, we used their results based on odometry without mapping. For ICP, GICP, and WGICP, we use a  voxel of size 0.5 for downsampling. For VGICP, we use a voxel of size 0.25 for downsampling, and 0.5 for the algorithm. Note that our WGICP exhibits low translational error for every sequence, including the test dataset; our method achieves {\bf 30\%} better translational accuracy than baseline GICP on the training dataset.}
\label{table:accuracy-comparison-wo-rejection}

\begin{threeparttable}

\begin{tabular}{c | cc | cc | cc | cc || cc | cc}
\hline
 & \multicolumn{2}{c|}{ICP~\cite{besl1992method}} & \multicolumn{2}{c|}{GICP~\cite{segal2009generalized}} & \multicolumn{2}{c|}{VGICP~\cite{koide2021voxelized}} &
 \multicolumn{2}{c||}{WGICP (Our)} &
 \multicolumn{2}{c|}{Velas et al.~\cite{velas}} & \multicolumn{2}{c}{LO-Net~\cite{li2019net}}  \\
\hline
Seq. & $T_{rel}$ & $R_{rel}$& $T_{rel}$ & $R_{rel}$& $T_{rel}$ & $R_{rel}$& $T_{rel}$ & $R_{rel}$& $T_{rel}$ & $R_{rel}$& $T_{rel}$ & $R_{rel}$  \\
\hline
00$\dagger$ & 9.44 & 4.07 & \textbf{1.44}\tnote{1} & 0.66 & 4.42 & \textbf{0.64}  & 1.80 & 1.03 & 3.02 & -\tnote{2} & 1.47 & 0.72 \\
\hline
01$\dagger$ & 63.69 & 12.43 & 5.14 & 1.06 & 64.54 & 1.68  & \textbf{2.16} & \textbf{0.69} & 4.44 & - & 1.36 & 0.47 \\
\hline
02$\dagger$ & 24.06 & 5.91 & \textbf{2.18} & \textbf{0.70} & 4.33 & 1.09  & 2.34 & 1.28 & 3.42 & - & 1.52 & 0.71 \\
\hline
03$\dagger$ & 22.62 & 4.92 & 1.80 & 0.91 & 1.49 & 0.88  & \textbf{1.64} & \textbf{0.75} & 4.94 & - & 1.03 & 0.66 \\
\hline
04$\dagger$ & 56.24 & 5.04 & 2.25 & 0.84 & 73.91 & \textbf{0.69}  & \textbf{0.47} & 0.90 & 1.77 & - & 0.51 & 0.65 \\
\hline
05$\dagger$ & 4.49 & 1.87 & \textbf{1.01} & \textbf{0.61} & 5.30 & 0.64  & 1.05 & 0.83 & 2.35 & - & 1.04 & 0.69 \\
\hline
06$\dagger$ & 6.13 & 1.41 & 0.81 & \textbf{0.40} & 3.81 & 0.48  & \textbf{0.48} & 0.68 & 1.88 & - & 0.71 & 0.50 \\
\hline
07$*$ & 5.66 & 1.90 & \textbf{0.96} & 0.55 & 3.01 & \textbf{0.40} & 1.44 & 0.79 & 1.77 & - & 1.70 & 0.89  \\
\hline
08$*$ & 11.07 & 4.49 & 1.78 & 0.95 & \textbf{1.59} & \textbf{0.57} & 1.75 & 1.09 & 2.89 & - & 2.12 & 0.77  \\
\hline
09$*$ & 24.00 & 6.76 & 1.42 & \textbf{0.55} & 4.45 & 0.87 & \textbf{1.29} & 0.95 & 4.94 & - & 1.37 & 0.58  \\
\hline
10$*$ & 19.20 & 5.69 & \textbf{1.26} & \textbf{0.66} & 3.07 & 0.68 & 1.74 & 1.03 & 3.27 & - & 1.80 & 0.93 \\
\hline
\hline
mean$\dagger$ & 26.67 & 5.09 & 2.09 & \textbf{0.74} & 22.54 & 0.87 & \textbf{1.42} & 0.88 & 3.12 & - & 1.09 & 0.63 \\
\hline
mean$*$ & 14.98 & 4.71 & \textbf{1.36} & 0.68 & 3.03 & \textbf{0.63} & 1.56 & 0.96 & 3.22 & - & 1.75 & 0.79 \\
\hline

\hline
\end{tabular}
\begin{tablenotes}
    \item [$\dagger$]: Training sequences used to train networks for WGICP.
    \item [$*$]: Test sequences used for evaluation.
    \item [$T_{rel}$]: Average translational error (\%) on length of 100m-800m, computed with KITTI. 
    \item [$R_{rel}$]: Average rotational error ($^{\circ}$/100m) on length of 100m-800m, computed with KITTI. 
    \item [1]: The lowest error among ICP-based methods are shown in bold face.
    \item [2]: Rotational errors are not available.
\end{tablenotes}
\end{threeparttable}
\vspace*{-0.5em}

\label{tab:RT_error}
\end{table*}

\begin{table*}[t]
\centering 
\caption{{\bf Odometry error comparison between GICP and WGICP in terms of rejecting outliers}. Voxel size (m) denotes the size of voxel used for preprocessing an input point cloud. While GICP uses all points generated from voxel downsampling, WGICP rejects some of them based on the computed weights. The portion of rejected points is denoted as Rejection (\%). The ratio of remaining points that are used by the algorithm is denoted as \# Point (\%). Note that for GICP, average error increases for every sequence when the number of points decrease. However,  WGICP is quite robust in such cases. The errors do not increase much even when we reject {\em 50\%} of the points. In many cases, the error decreases with outliers rejection.}
\label{table:accuracy-comparison-w-rejection}
\begin{threeparttable}
\begin{tabular}{c | cc | cc | cc | cc | cc | cc | cc | cc}
 & \multicolumn{4}{c|}{GICP} & \multicolumn{12}{c}{WGICP} \\
\hline
Voxel Size ($m$) & \multicolumn{2}{c|}{1.0} & \multicolumn{2}{c|}{2.0} & \multicolumn{6}{c|}{1.0} & \multicolumn{6}{c}{2.0} \\
\hline
Rejection (\%) & \multicolumn{4}{c|}{0} & \multicolumn{2}{c|}{25} & \multicolumn{2}{c|}{50} & \multicolumn{2}{c|}{75} & \multicolumn{2}{c|}{25} & \multicolumn{2}{c|}{50} & \multicolumn{2}{c}{75}  \\
\hline
\# Point (\%) & \multicolumn{2}{c|}{5.2} & \multicolumn{2}{c|}{2.3} & \multicolumn{2}{c|}{3.3} & \multicolumn{2}{c|}{2.6} & \multicolumn{2}{c|}{1.3} & \multicolumn{2}{c|}{1.8} & \multicolumn{2}{c|}{1.2} & \multicolumn{2}{c}{0.6} \\
\hline
\hline
Seq. & $T_{rel}$ & $R_{rel}$ & $T_{rel}$ & $R_{rel}$ & $T_{rel}$ & $R_{rel}$ & $T_{rel}$ & $R_{rel}$ & $T_{rel}$ & $R_{rel}$ & $T_{rel}$ & $R_{rel}$ & $T_{rel}$ & $R_{rel}$ & $T_{rel}$ & $R_{rel}$ \\
\hline
00$\dagger$ & 2.1 & 1.0 & 4.1 & 1.8 & 2.4 & 1.1 & 2.5 & 1.3 & 4.6 & 2.4 & \textbf{4.1} & 2.2 & 5.5 & 2.6 & 8.8 & 5.2 \\
\hline
01$\dagger$ & 3.4 & 0.6 & 9.2 & 1.8 & \textbf{3.3} & 0.7 & 8.1 & 1.8 & 28.7 & 9.1 & \textbf{7.3} & \textbf{1.5} & \textbf{6.7} & \textbf{1.6} & 14.7 & 4.0 \\
\hline
02$\dagger$ & 3.4 & 1.1 & 6.6 & 2.1 & \textbf{2.5} & \textbf{1.0} & \textbf{2.8} & 1.3 & 3.8 & 1.6 & \textbf{5.6} & \textbf{2.1} & \textbf{5.2} & 2.2 & 6.9 & 3.2 \\
\hline
03$\dagger$ & 3.7 & 1.0 & 7.1 & 1.8 & 4.3 & 1.3 & 6.2 & 1.7 & 4.7 & 1.5 & \textbf{4.7} & \textbf{1.4} & \textbf{5.7} & \textbf{1.4} & \textbf{6.5} & 1.9 \\
\hline
04$\dagger$ & 3.0 & 0.7 & 9.7 & 0.9 & 3.1 & 0.8 & \textbf{2.3} & 0.8 & 26.1 & 7.7 & \textbf{5.5} & 1.3 & \textbf{2.3} & 2.6 & \textbf{7.9} & 4.5\\
\hline
05$\dagger$ & 1.8 & 0.8 & 3.8 & 1.5 & \textbf{1.8} & \textbf{0.8} & \textbf{1.5} & \textbf{0.7} & 4.0 & 1.6 & \textbf{3.6} & 1.5 & 4.1 & 2.8 & 8.3 & 3.5 \\
\hline
06$\dagger$ & 0.7 & 0.4 & 2.6 & 0.8 & 1.9 & \textbf{0.3} & 1.8 & 0.7 & 4.9 & 1.8 & 3.0 & 1.0 & \textbf{2.1} & 1.0 & 3.2 & 1.9\\
\hline
07$*$ & 3.0 & 1.4 & 5.9 & 2.5 & \textbf{2.5} & \textbf{1.1} & \textbf{2.6} & 1.4 & \textbf{2.2} & 1.6 & 8.0 & 3.4 & 10.0 & 4.7 & 11.4 & 7.0\\
\hline
08$*$ & 2.9 & 1.4 & 5.6 & 2.3 & \textbf{2.1} & \textbf{1.0} & \textbf{2.6} & \textbf{1.1} & 3.1 & 1.5 & \textbf{5.1} & \textbf{2.3} & \textbf{3.9} & \textbf{2.1} & 6.9 & 4.0 \\
\hline
09$*$ & 1.8 & 0.7 & 4.8 & 1.5 & 2.5 & 1.1 & 3.0 & 1.3 & 4.8 & 2.4 & \textbf{3.6} & \textbf{1.4} & \textbf{3.8} & 1.9 & 6.4 & 3.8\\
\hline
10$*$ & 2.3 & 0.7 & 4.9 & 1.9 & 2.9 & 1.1 & 3.4 & 1.7 & 6.1 & 3.2 & \textbf{3.9} & \textbf{1.5} & \textbf{4.3} & \textbf{1.8} & 14.9 & 5.8\\
\hline
mean & 2.6 & 1.0 & 5.3 & 1.9 & \textbf{2.4} & \textbf{0.9} & 2.8 & 1.2 & 5.2 & 2.2 & \textbf{4.8} & 1.9 & \textbf{4.8} & 2.3 & 8.0 & 4.0\\

\hline
\end{tabular}
\begin{tablenotes}
    \item [1]: Errors that are less than those from corresponding baseline GICP are rendered in bold face.
\end{tablenotes}
\end{threeparttable}
\vspace*{-1.0em}
\end{table*}

%% file: 5_experiments.tex
\section{Implementation and Results}
\label{sec:experiments}

In this section, we describe our implementation and compare its performance with prior odometry algorithms. Furthermore, we also use our approach for SLAM tasks by combining with loop closure~\cite{cattaneo2022lcdnet}. We use the KITTI dataset~\cite{Geiger2013IJRR} for all of our evaluations.

Our algorithm is implemented in Python and PyTorch 1.9~\cite{paszke2019pytorch}. All evaluations were run on an Intel\textregistered~Xeon\textregistered~W-2255 CPU @ 3.70GHz, and our networks were trained on an NVIDIA RTX A5000 GPU. We use~\cite{koide2021voxelized} to measure our running time for evaluation.

\subsection{Performance Without Rejecting Outliers}

We observe that our trained network generally achieves better odometry results in terms of translational error than other algorithms. Note that since we use the loss term in Equation~\ref{eq:loss}, our network is trained to reduce the translational error as compared to the rotational error.

In Table~\ref{table:accuracy-comparison-wo-rejection}, we observe that our WGICP method generally achieves low translational error for every sequence. We also observe that our method gives more stable results than other ICP-based methods, as its maximum translation error is $2.34$, while that of GICP is $5.14$. Even though LO-Net~\cite{li2019net} provides better results for some  training dataset, it does not generalize well to test datasets. This might be due to overfitting. Even though our method gives slightly larger translational error than baseline GICP for the test dataset, we observe that the error is still lower than that of LO-Net and is stable. Even without using rejection outliers,  we observe stable results without any overhead.


\subsection{Performance With Rejecting Outliers} 

\begin{figure}[h!]
    \centering
    \includegraphics[width=0.7\linewidth]{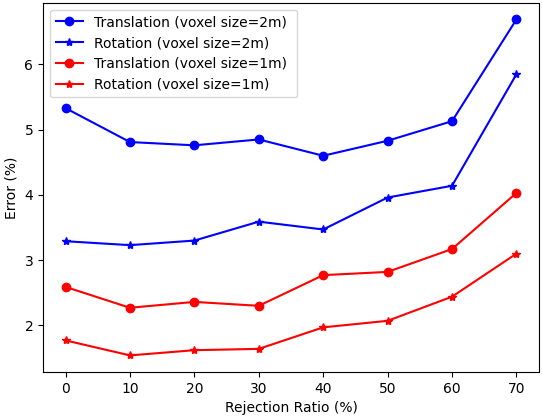}
    \caption{{\bf Error plots with different rejection ratios}. The blue lines show the translation and rotation errors with an increasing rejection ratio from the voxels with size of 1m. The red lines indicate the errors from the voxels with size of 2m. 
    Note that errors decrease as we reject  the outliers first. Later it increases, after we reject more than 60\% of points.}
    \label{fig:pointsrejection}
\end{figure}

By rejecting the outliers, not only can WGICP offer improved performance over the GICP algorithm, but it can also improve its accuracy. In Table~\ref{table:accuracy-comparison-w-rejection}, we observe that the error increases when the number of points decrease for GICP. However, since our WGICP rejects outliers that hinder correct pose estimation, we can observe that error decreases even when we reduce the number of points using rejection outliers. In some cases, we can observe that the error is less than the baseline error even when we reject {\em 75\%} of the input point clouds. We observe the same behavior in Figure~\ref{fig:pointsrejection}. When we reject {\em 10\%-30\%} of input points, error decreases because we remove redundant points that can increase the error in pose estimation. Note that even when we reject more points (i.e., up to {\em 50-60\%}), the error does not increase considerably. Since we can accelerate the overall algorithm by using a reduced number of points (Figure~\ref{fig:comp-time-rejection}), we can improve both the accuracy and speed with our method. To be specific, our algorithm runs {\bf 2}$\times$ faster than GICP with similar results, or  provides {\bf 10\%} higher accuracy than GICP with {\bf 10\%} less in running time.

\subsection{Application to SLAM}
\begin{figure}
    \centering
    \includegraphics[width=0.4\linewidth,height=0.5\linewidth]{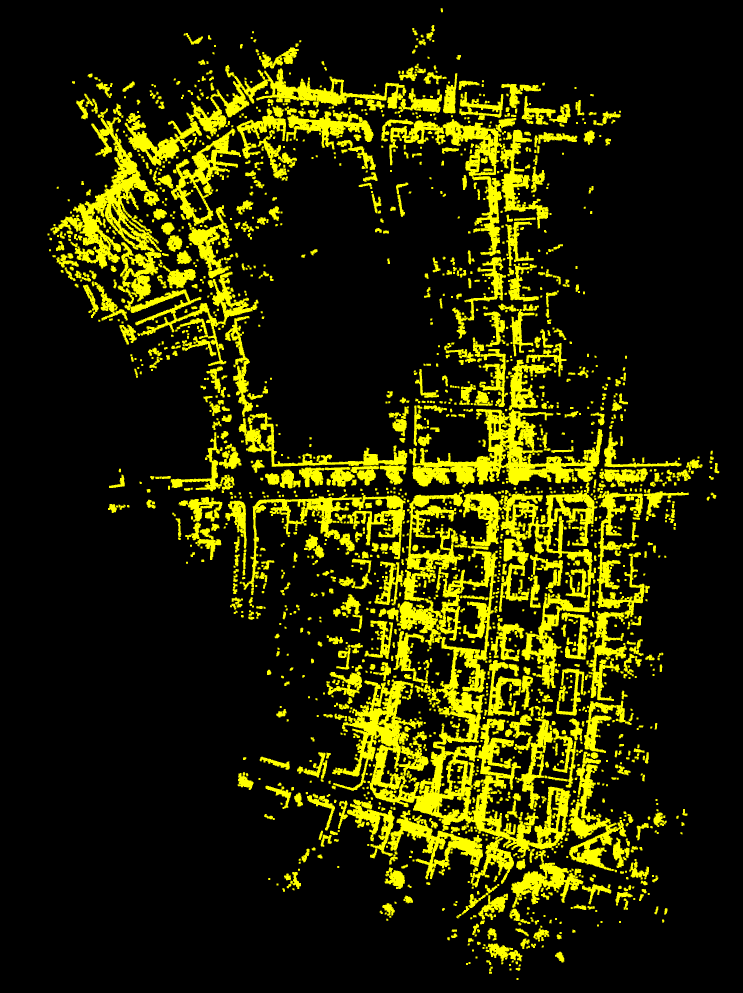}
    \includegraphics[width=0.45\linewidth,height=0.52\linewidth]{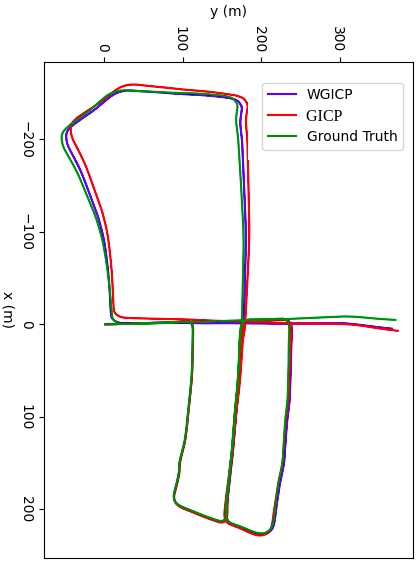}
    \caption{{\bf Differentiable odometry on the KITTI dataset for mapping and SLAM}. The maps are generated from the KITTI Sequences 05.  RIGHT:  the trajectories of our approach WGCIP (purple), GICP(red) and the ground truth (green). From the top left and right sides of the trajectories, we observe that our approach generates closer trajectory to the ground truth than GICP.
    }
    \label{fig:mapping}
    \vspace{-5mm}
\end{figure}

Our approach can also be used to improve the performance of SLAM systems. We implement the LCD loop closure~\cite{cattaneo2022lcdnet} method and combine with our odometry algorithm and use it to compute an accurate map. With the Kitti dataset sequence 5, we compare the performance of GICP and our approach in terms of mapping, as shown in Figure \ref{fig:mapping}. The figures show the w rectified map. Our approach (WGICP) is closer to the ground truth trajectory, where the average error of the trajectory is $0.00016$ m. The GICP approach exhibits worse performance, where the trajectory deviates from the ground truth and the average error is $0.000275$ m. Please refer to our project WWW for more comparisons and analysis.

\input{5-3-figure}

%% file: 5-3-figure.tex
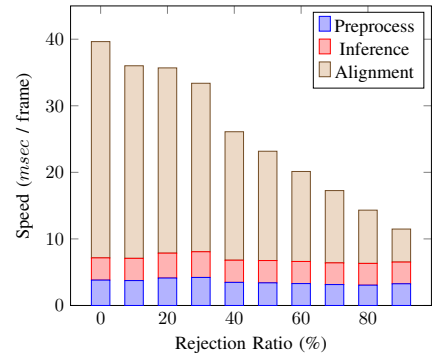
\begin{figure}
    \centering

    \begin{tikzpicture}[scale=0.70]

\begin{axis}[
    ybar stacked,
    xlabel=Rejection Ratio (\%),
    ylabel=Speed ($msec$ / frame),
    ymin = 0, ymax=45,
    ]

\addplot coordinates {
    (90, 3.26)
    (80, 3.05)
    (70, 3.14)
    (60, 3.29)
    (50, 3.40)
    (40, 3.47)
    (30, 4.21)
    (20, 4.12)
    (10, 3.74)
    (0, 3.81)
};
\addlegendentry{Preprocess}

\addplot coordinates {
    (90, 3.26)
    (80, 3.27)
    (70, 3.26)
    (60, 3.32)
    (50, 3.35)
    (40, 3.33)
    (30, 3.86)
    (20, 3.74)
    (10, 3.34)
    (0, 3.34)
};
\addlegendentry{Inference}

\addplot coordinates {
    (90, 4.96)
    (80, 8.00)
    (70, 10.86)
    (60, 13.52)
    (50, 16.42)
    (40, 19.30)
    (30, 25.31)
    (20, 27.83)
    (10, 28.93)
    (0, 32.49)
};
\addlegendentry{Alignment}

\end{axis}
\end{tikzpicture}
    
    \caption{{\bf Average computation time of WGICP based on the ratio of rejecting outliers}. We measure the performance on a single-core CPU with voxel-based downsampling (size of 1m) and alignment, and use GPU for inferencing the point weights. Note that the computation time largely depends on the number of points used by the GICP algorithm; the inference time ($\le 4ms$) is negligible by comparison.}
    \label{fig:comp-time-rejection}
    \vspace{-3mm}
\end{figure}

%% file: 6_conclusion.tex
\section{CONCLUSION AND LIMITATIONS}
\label{sec:conclusions}

We present a fully {\em differentiable weighted generalized iterative closest point} (WGICP) approach for lidar odometry. Our approach is capable of selecting points with higher weights to calculate the transformation matrix between two frames of 3D points. Compared with prior, our approach achieves faster performance (up to {\bf 2X} improvement) and better accuracy (up to {\bf 30\%}) than existing GICP methods. We also demonstrate the benefits in terms of generating more accurate maps.

Our approach has some limitations. First, we use a uniform rejection ratio for every frame, but the experimental results show that the optimal rejection ratio differs by frame. It may be possible to further improve both the accuracy and speed by changing the rejection ratio in an adaptive manner. Moreover, it might be possible to exploit the per-point feature vectors generated by PointNet. Since such feature vectors are more descriptive than scalar weight values, we might be able to use them for finer matching. It would also be useful to generalize this approach to other domains. Our learning based method is complimentary to state of the art methods like CT-GICP~\cite{dellenbach2022ct}. It may be useful to combine such approaches to further improve the performance.